# Clinical Trial Recommendations Using Semantics-Based Inductive Inference and Knowledge Graph Embeddings

Murthy V. Devarakonda,* Smita Mohanty, Raja Rao Sunkishala, Nag Mallampalli, and Xiong Liu

Biological Research, Novartis, Cambridge, MA, USA.

**Abstract:** Designing a new clinical trial entails many decisions, such as defining a cohort and setting the study objectives to name a few, and therefore can benefit from recommendations based on exhaustive mining of past clinical trial records. Here, we propose a novel recommendation methodology, based on neural embeddings trained on a first-of-a-kind knowledge graph of clinical trials. We addressed several important research questions in this context, including designing a knowledge graph (KG) for clinical trial data, effectiveness of various KG embedding (KGE) methods for it, a novel inductive inference using KGE, and its use in generating recommendations for clinical trial design. We used publicly available data from *clinicaltrials.gov* for the study. Results show that our recommendations approach achieves relevance scores of 70%-83%, measured as the text similarity to actual clinical trial elements, and the most relevant recommendation can be found near the top of list. Our study also suggests potential improvement in training KGE using node semantics.

## 1 INTRODUCTION

Clinical trials are essential to development of new medical treatments and therapies, as they involve systematic study of interventions to determine their efficacy and safety on participants. The conceptual stage of designing a clinical trial (Scott 2010), which is an early stage of the process, involves defining the endpoints (objectives) of the study, identifying eligible participants, designing the study protocol, outcome measurement methodology, and ensuring regulatory compliance. The goal is to ensure that the study is framed effectively to obtain results in a timely manner. As such, the designers can benefit from understanding historical trial designs for similar indications and interventions. This historical perspective can be provided as relevant recommendations for a given clinical trial design. Here we hypothesis that trained neural embeddings of a knowledge graph representing clinical trials is a sound basis to generate such recommendations.

ClinicalTrials.gov (abbreviated as CT.gov in the rest of the paper), maintained by the United States National Library of Medicine (NLM), is a registry and results database of publicly and privately supported clinical studies of human participants conducted around the world. In the USA, clinical trials are required to be registered in the database and are expected to be updated regularly. As of March 2023, it contains information on over 400,000 clinical studies from around the world. CT.gov provides a query capability to find relevant trials, and in addition, vendors and public databases also regularly extract the data and provide query capabilities on the extracted data. However, a recommendation engine, that can provide suggestions for design elements for a trial being designed, is lacking.

An objective of this study is to support the type of recommendations generation shown in Figure 1. Given a draft descriptive title of a new clinical trial, the objective is to generate recommendations most useful for the design of the trial. In the Figure, an example draft title and ideal recommendations expected from the system for the primary endpoint and inclusion criteria. are shown (which were taken from an actual clinical trial recorded in CT.gov).

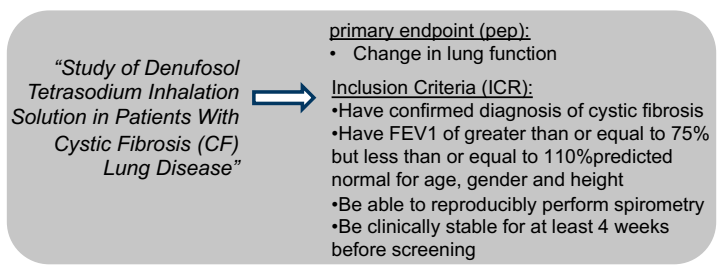

*Figure 1. An example of expected clinical trial design recommendation. The input is the title of the proposed clinical trial as shown on the left and the system is expected to return recommendations shown on the right.*

Another motivation for this work is the methodological exploration of knowledge graphs for biomedical data. Here we propose a first-of-a-kind knowledge graph for clinical trials design. While *transductive* inference for knowledge graphs of biological data has been studied previously for predicting missing relations, for example, as for drug repurposing in (Zheng et al. 2021), *inductive* inference (Teru, Denis, and Hamilton 2020) has not been studied in biomedical knowledge graphs. However, in the general domain, inductive inference has received some attention (S. Wang et al. 2023; Li et al. 2023). Here, we provide a unique datapoint for inductive inference in biomedical knowledge graphs. Furthermore, the impact of graph structure, such as the graph density, on representation and inference has been explored rarely in any domain. We consider these methodological issues in the context of the clinical trial data.

To conduct the studies outlined above, we propose NVKG, **No**vartis **K**nowledge **G**raph, for representing clinical trials data for design recommendations and predictions, and we evaluate various approaches for training knowledge graph embeddings (KGE) and the methodology for transductive and inductive inference using the embeddings. We evaluated KGE methods, ranging from the well-established node2vec and TransE methods to the latest graph neural networks such as the Graph Attention Networks. We compared our transductive inference results with those obtained in PharmKG (Zheng et al. 2021), which is one of the most relevant and recent KG for biomedical data, to consider the influence of graph characteristics on the performance of various models.

We developed a novel inductive inference technique for downstream tasks using the knowledge graph embedding, viz., making recommendations for two key clinical trial design elements: endpoints and eligibility criteria. In clinical trials, endpoints are the systematically measured objectives of the trial and thus represent the effectiveness of the proposed medical intervention. Inclusion and exclusion criteria are the rules that determine whether an individual is eligible to participate in a clinical trial. Our goal is to help conceptual clinical trial designers by providing the most relevant recommendations. We used a "blind set" of historical trials (not used in building the KG) to evaluate the inductive inference technique. Thus, contributions of our study are:

- First-of-a-kind knowledge graph to represent clinical trials data for design recommendations and predictions.
- Evaluation of multiple KGE training methods of the clinical trials KG based on transductive inference.
- Novel semantics-based inductive inference technique for generating KG embeddings for nodes unseen in training.
- Demonstration of how the inductive inference can be used for providing relevant recommendations.

The paper is organized as follows. The next section presents related work, which is followed by methods we employed for constructing the knowledge graph, KGE training, and inductive inference as well as design of experiments and performance evaluation metrics. Results are presented in the subsequent section, which is followed by discussion and concluding sections.

## 2  RELATAED WORK

Two recent studies of biomedical knowledge graphs are most relevant to our work. The first is PharmKG (Zheng et al. 2021), which proposed a knowledge graph relating disease, gene, and chemicals using structured databases as well as mined results from biomedical literature. While the methods used to train KGE and the transductive inference methodology in PharmKG are relevant to our study, the data represented in PharmKG is very different from ours. PharmKG is characterized by many relation types among just three node types whereas our graph is quite the opposite. Furthermore, PharmKG did not explore inductive inference.

The second closely related study is CTKG (Chen et al. 2022), which proposed a knowledge graph to represent clinical trials that have outcomes (a relatively small portion of the clinical trials in CT.gov) and the goal was to use trained graph embeddings for drug re-purposing. The key distinction with our work is that CTKG focused on representing outcomes whereas NVKG is designed to represent (and then to recommend or predict) clinical trial design elements such as the endpoints and eligibility criteria. Furthermore, CTKG has only considered TransE for KGE whereas we have assessed several methods, including the latest graph neural networks.

Beyond these specific studies, related research broadly falls into three major areas. First, structuring biomedical data as graphs, second developing graph embeddings, and lastly downstream applications of the embeddings. In the following paragraphs, we will discuss the broader related research in these areas to position our approach appropriately. A comprehensive review of the state of the art can be found in (Abu-Salih et al. 2023).

### 2.1  Knowledge Graphs in the Biomedical Domain

Knowledge graphs have been increasingly used in biomedical research and vendor products in recent years, due to their ability to integrate diverse data sources, facilitate neural representation of the knowledge, and enable data-driven discovery through inference. Knowledge graph representation grew out of high quality, often manually curated, structured databases of biomedical knowledge, such as Drug Bank (Wishart et al. 2018), Therapeutic Target Database (TTD) (Y. Wang et al. 2020), Online Mendelian Inheritance in Man (OMIM) (Hamosh et al. 2000), and PharmGKB (Hewett et al. 2002). Belleau et al applied semantic web technologies to create RDFs from public bioinformatics databases (Belleau et al. 2008), which may be the beginning of knowledge graphs development for biomedical data.

There were also efforts to mine large biomedical literature with unsupervised techniques (Percha and Altman 2018; Hong et al. 2020) and use the extracted relations to build knowledge graphs which in turn were used to generate drug re-purposing suggestions. Several other studies (Ou et al. 2016; Alshahrani et al. 2017; Breit et al. 2020; Mohamed, Nováček, and Nounu 2020; Sosa et al. 2020) also built knowledge graphs from scientific publications and used them for drug repurposing studies.

Our goal here was to develop a structural representation of clinical trials data for enabling downstream applications to help clinical trial design, we considered relevant modeling issues such as what information about a clinical trial should be represented? What should be considered as a node and what should be modeled as its attribute(s)? Our KG design provides a useful example of how to address such design issues.

## 2.2 Representations of Biomedical Knowledge Graphs

While the early experiments with knowledge graphs used graph algorithms for reasoning (and therefore did not benefit from data-driven model training), the modern approach is to train low-dimensional neural embeddings of knowledge graphs (KGEs). These embeddings enable prediction of missing relationships which help to *complete* the knowledge. Many excellent surveys of KGE methods exist (Cao et al. 2022; Dai, Wang, and Xiong 2020; Q. Wang et al. 2017) and so here we review only the methods we used in our study. Broadly the approaches may be characterized as translational, semantics-based, and neural network based.

The translational approach, pioneered by TransE (Bordes et al. 2013) and subsequently followed by a series of distinct advances resulting in TransR (Lin et al. 2015), models a relation $r$ in a triplet $(h, r, t)$ as a translation of the head entity $h$ to the tail entity $t$ in the vector space such that $h + r \sim t$. While TransE uses translation in a single latent space and is therefore challenged when 1-N, N-1, and N-M relations exist, TransR improves representation in those cases by using different latent spaces for entities and relations. CTKG has used only TransE but our study as well as PharmKG used TransE as the baseline for evaluating multiple other approaches, including TransR.

We should note that node2vec (Grover and Leskovec 2016), an approach based on using word2vec to sequences of nodes, can be thought of as a special case of the translational approach. In node2vec, sequences of random walks are generated from a graph and these sequences are used to train embeddings for nodes just as word embeddings are trained in wod2vec using the SkipGram or CBOW contextual modeling. Despite its simplicity and limitations in representing N-M relations, node2vec, can represent contextual meanings well when 1-1 relations are dominant. Therefore, we also evaluated this approach for training embeddings of our knowledge graph.

The semantics-based models leverage latent-semantics of entities and relations. For example, RESCAL (Nickel, Tresp, and Kriegel 2011) formalizes the notion that entities are similar if they are related to similar entities via the same relationships. DistMult (Yang et al. 2015) and ComplEx (Trouillon et al. 2016) advanced the notion through optimized models for representing semantic similarities in the graph structure. We evaluated ComplEx for our graph embeddings.

Graph neural networks (GNNs) (Zhou et al. 2020) are now the *de facto* technique for training graph embeddings, given the success of neural networks in analyzing other data formats such as images and natural language text. Variations exist, but fundamentally GNNs use message passing to train the representation of a node as an *aggregation* of its neighbors' representations. This message passing architecture is encapsulated in an encoder-decoder network, where the encoder generates representations for nodes using message passing and decoder attempts to reproduce the graph from the embeddings. Large number of epochs are conducted, updating representations in each epoch, thus converging to capture the global structure while making use of the local structure. The aggregation may be achieved using a CNN, Transformer, or any other deep learning architecture.

R-GCN (Schlichtkrull et al. 2018), ConvE (Dettmers et al. 2018), and ConvKB (Nguyen et al. 2018) are GNNs that used convolutional neural network as the aggregator functions. GAT (Velickovic et al. 2018) is an attention-based messaging passing graph neural networks. PharmKG adapted GAT as HRGAT and compared its performance against ConvKB along with other models. We chose ConvKB and HRGAT to conduct experiments for our KG as they represent the state of the art GNNs and readily usable implementations exist.

## 2.3 Downstream Applications for Biomedical Knowledge Graphs

Lastly, downstream applications are the key value to be derived from developing a KG and training embeddings for it. While the design of the KG schema and the data used to populate the KG address unique needs specific to each application, several common themes are apparent from the prior studies. For example, link prediction is the common task to identify missing relations, which leads to hypothesis generation that may hold the key to a new discovery. Another aspect is the power of integration. As multiple databases or data from multiple sources are integrated into a single knowledge graph, the integrated knowledge graph may yield insights that are not evident otherwise. Below we describe a few downstream applications from previous research that are relevant to our study.

- **Drug re-purposing**: Drug re-purposing is recurring theme with many KG applications. The Drug Repurposing Hub[1] identified potential new uses for existing drugs by integrating data from multiple sources, including gene expression data, drug databases, and literature. PharmKG and CTKG have also been developed with drug re-purposing as a goal.
- **Clinical decision support**: Knowledge graphs have been used to support clinical decision-making by integrating patient data, medical knowledge, and clinical guidelines (Santos et al. 2022). For example, the OHDSI Knowledge Graph[2] is a comprehensive ontology of medical concepts and relationships that can be used to standardize and harmonize electronic health records for clinical research.
- **Biomedical knowledge discovery**: Knowledge graphs have been used to discover new relationships between biological entities (new "knowledge") and to generate new hypotheses for further investigation (Zeng et al. 2022) (Ren et al. 2022). For example, the Bioteque project (Fernández-Torras et al. 2022) has developed a knowledge graph that integrates 12 biological entities, including genes, cells, tissues, diseases, and pathways, to characterize and predict a wide range of biological observations.
- **Precision medicine**: Knowledge graphs have been used to support personalized medicine by integrating patient data, genomic data, and medical knowledge to predict treatment outcomes and identify potential therapies. For example, the Knowledge Graphs for Precision Medicine project (Chandak, Huang, and Zitnik 2023) is developing a knowledge graph that integrates genomic data, clinical data, and medical knowledge to support cancer treatment decision-making.

While knowledge graphs have been proposed for various downstream applications as reviewed above, NVKG is the first to use this approach to clinical trials data for recommending design elements or predicting completion probabilities.

## 2.4 Transductive vs Inductive Inference

Many useful knowledge graphs are built with the assumption that the data represented in the KG forms a closed world, i.e., no new nodes (types or instances) and that new relation types will not be added. In such a setting, the trained model can be used to fill missing knowledge gaps. For example, in PharmKG, trained embeddings were used to hypothesize a treatment relationship between Etoposide and Alzheimer Disease even though the input to the graph did not have a relation between them. This kind of inference is known as Transduction or Transductive inference.

In real life, new nodes are often added to a knowledge base. For example, at the early stages of the coronavirus pandemic, it would have been necessary to introduce Covid-19 as a new disease node into a graph of disease-treatment-target. The process would include connecting the new node to the other nodes (which may also be new or existing) in the graph using known real world knowledge about the new node vis-à-vis existing nodes. To reason about the new node in the latent space of KGE, it is necessary to train or generate embeddings for the new node. It would be onerous to retraining the whole KG for this purpose and will not be feasible if new nodes are added frequently. The scenario is like handling of the out-of-vocabulary (OOV) words in NLP. New approaches are being developed for the purpose, for example, NodePiece (Galkin et al. 2022) trains embeddings for a node based on its structural pattern among its neighbors and thus can be used to generate embeddings for a new node, after certain number of relationships (edges) are established to existing nodes, without retraining the entire KG.

---

[1] https://www.broadinstitute.org/drug-repurposing-hub

[2] https://www.ohdsi.org/2022showcase-43

We first trained and tested NVKG using the transductive approach. However, to address the needs of the intended downstream applications, we developed a novel semantics-based inductive approach to determine embeddings for new nodes without retraining the entire KG. The new node embeddings, in conjunction with embeddings of the existing nodes, is used to generate recommendations and predictions. We provide more details of the approach in the Methods section below.

## 3 METHODS

The methods we employed here spans multiple areas and we describe each of them in turn, starting with the schema design and data used in knowledge graph construction. Next, we discuss KGE training techniques and the metrics for evaluation of the embeddings. Subsequently, we describe the semantics-based technique for generating embeddings for a new node and its application in the downstream tasks.

### 3.1 Knowledge Graph Construction

In developing NVKG we used data from three sources: the clinical trials repository (CT.gov as of August 2022), data about drugs (targets and mechanism of action) from DrugBank as curated in TrialTrove, and UMLS with unique ids for entities such as indications, interventions, targets, and mechanism of action. NVKG schema was designed based on the semantics of the data we used from these sources.

In designing the NVKG schema, the first challenge was how to represent a text entity, such as an inclusion criterion, in NVKG. Since text entities are most common in NVKG, they represent the most important data. For example, when we want to represent two inclusion criteria, from two clinical trials, that are textually similar we may do so, as shown in Figure 2, with four nodes and three relations (option 1) or simply with one node (as in option 2).

Consider the options in representing two inclusion criteria from two trials. The first option uses a node for each inclusion criterion (ICR nodes) and a pair of separate text nodes, Eligibility Criterion Text (ECT) nodes. The text of an entity is stored as the attribute of an ECT node. ICR nodes are connected to ECT nodes using "has_text" relation and a "similar" type edge connects the two ECT nodes if the wording of the two criteria is similar (based on a text similarity score). This schema well captures the trial data and potential similarities among their textual design elements.

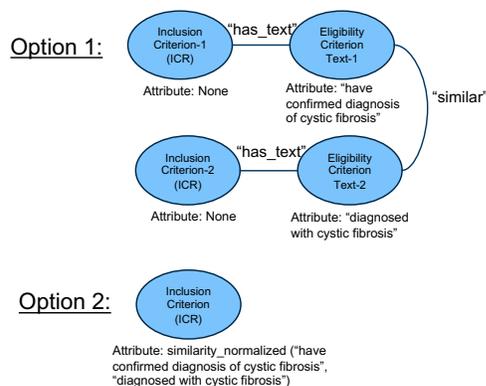

*Figure 2. Illustration of design choices in representing an inclusion criterion, for example, "Have confirmed diagnosis of cystic fibrosis" in the knowledge graph*

However, experiments showed that option 1 resulted in many (3-4 times as many) additional nodes and edges and did not improve entity embeddings as assessed by link prediction. Therefore, we chose option 2, where we used text normalization based on similarity and the normalized text was stored as the attribute of the entity's node (Inclusion Criteria node in Figure 2). With this approach, the two clinical trials are connected to the same inclusion criteria node if they had similarly worded inclusion criterion. To perform text normalization, we first clustered textual values of entities (using *fasttext* similarity scores and clustering by entity types, e.g., eligibility criteria separately from endpoints) and then used the textual values of the cluster centroids as the normalized values.

Note that since clinical trials are not directly related to each other in the data source (CT.gov), the only way KG representations learn trial similarity is through common indications, interventions, eligibility criteria, endpoints, and such design elements. Certain high frequency nodes (which tend to have categorical values) such as Phase and Gender do not provide specificity in terms of trial similarity as they commonly occur in many trials. Figure 3 shows the final NVKG schema and Table 1 shows the number of node and edge types in NVKG. The table also compares our KG statistics with CTKG and PharmKG. The ratio of edge types to node types (N/E ratio) in NVKG is like CTKG, but much lower than PharmKG's, which has far fewer node types but substantially larger number of edge types.

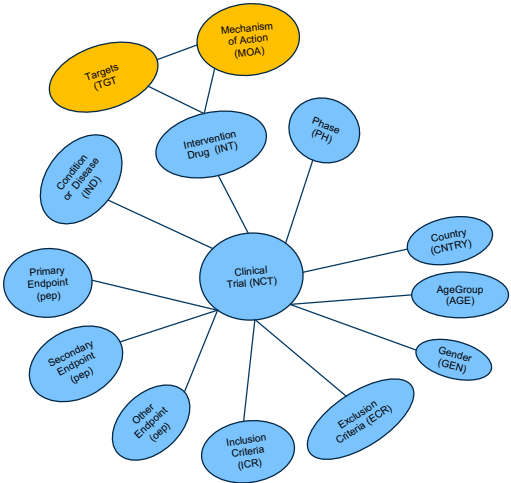

*Figure 3. The NVKG schema. Trials are connected not only through common nodes such as AgeGroup, Gender, Intervention, and Disease but also through textually similar end points and eligibility criteria.*

*Table 1. NVKG schema statistics and comparison with relevant biomedical KGs cited in the text. (\*node and edge statistics of the transductive study only)*

| Schema Objects | NVKG* | CTKG | PharmKG |
|---|---|---|---|
| Node types | 14 | 18 | 3 |
| Edge types | 15 | 21 | 29 |
| Nodes | 72,522 | 1,496,684 | 7,599 |
| Edges | 100,716 | 3,667,750 | 345,744 |
| N/E ratio | 1.389 | 2.451 | 45.50 |
| Num. of Trials | 2,713 | 9,680 | NA |

We filtered clinical trials in two ways before including their data in NVKG: First, only interventional studies were included (thus eliminating observational studies) and second, only drug-intervention trials were included. The rationale was to develop a KG for interventional drug studies only. The total number of trials in CT.gov that matched these filters were about 144,000 in August 2022. We built a KG using data from these trials and with the schema described above. However, experimenting this large graph required excessive GPU resources and time, and so we randomly selected a subset of 3201 trials from 11 disease areas, as shown in Table 2. The number of nodes and relations from the KG built from this subset is shown in Tables 3a and 3b. Most nodes represent eligibility criteria and end points, and most edges represent their relations to clinical trials. We note that nodes were standardized (i.e. ids were determined) in one of the following ways: Categorical values were used for such nodes as Phase (PH) and Age group (AGE), ids obtained from CT.gov were used for clinical trials nodes (NCT), UMLS concept unique identifiers (CUIs) were used for such nodes as interventions, indications, targets, and mechanism of action, and the MD5 sum of the normalized text (as described earlier) were used for textual nodes such as primary end point (pep) and inclusion criteria (ICR). When there is no coverage for an entity in UMLS we used the MD5 sum of the entity name as the identifier. In the KG, since ids uniquely identify nodes, id equality ensures that two different source nodes are connected to the same target node if the trials data implied such sharing.

*Table 2. Trials selected by disease area for NVKG experiments.*

| Disease Area | Num. of Trials |
|---|---|
| Asthma | 125 |
| Cystic Fibrosis | 440 |
| COPD (Chronic Obstructive Pulmonary Disease) | 183 |
| Early Rheumatoid Arthritis | 159 |
| Malignant Pleural Mesothelioma | 362 |
| Non-Small Cell Lung Cancer | 694 |
| Pulmonary Hypertension | 416 |
| Systemic Lupus Erythematosus | 452 |
| Tuberculosis | 370 |
| Total | 3201 |

*Table 3. (a) node and (b) edge counts in NVKG. Expansions for the abbreviations are in Figure 3.*

| Node type | Count |
|---|---|
| AGE | 6 |
| CNT | 114 |
| ECR | 24349 |
| GEN | 3 |
| ICR | 17050 |
| IND | 1035 |
| INT | 1220 |
| MOA | 414 |
| NCT | 2713 |
| PH | 8 |
| STA | 9 |
| TGT | 339 |
| oep | 792 |
| pep | 3997 |
| sep | 12229 |
| Total | 72,522 |

| Edge type | Count |
|---|---|
| INT:MOA | 1005 |
| INT:TGT | 753 |
| MOA:TGT | 2156 |
| NCT:AGE | 2713 |
| NCT:CNT | 7865 |
| NCT:ECR | 28378 |
| NCT:GEN | 2708 |
| NCT:ICR | 20074 |
| NCT:IND | 5100 |
| NCT:INT | 4818 |
| NCT:PH | 2713 |
| NCT:ST | 2713 |
| NCT:oep | 807 |
| NCT:pep | 4581 |
| NCT:sep | 14332 |
| Total | 100,716 |

In terms of NVKG's node attribute values, to clarify, nodes that represent categorical values, such as the AGE group or Intervention, do not have any attributes. Their values are encoded in their ids. However, textual nodes such as a clinical trial node or a primary endpoint node have a single attribute, i.e., the entity's text.

### 3.2 Training Knowledge Graph Embeddings

We trained knowledge graph embeddings using the following methods: *PecanPy* (Liu and Krishnan 2021) implementation of Node2Vec; the *pykeen* package (Ali et al. 2021) (Ali et al. 2022) implementations of TransE, TransR, ComplEx, and ConvKB; the code repository provided by the authors of PharmKG for HRGAT. For the purposes of transductive inference (link prediction), a stratified random split (i.e., proportional random samples from the disease areas) of the NVKG experimental dataset was used -- training (80%), validation (5%), and test (15%). PecanPy did not use the validation set and therefore it was merged with the training set. PecanPy parameters were 50 random walks per node and 1000 epochs on the random walks. For all the other models, we used the default settings in pykeen. We used 100 dimensions for embeddings and 1000 or 1500 epochs training as recommended by the authors of the code. Node and edge embeddings were initialized to random values for all except ConvKB and HRGAT. As recommended by the model developers, we used TransE generated embeddings as the initial values for them. All models generated embeddings for node instances and edge types except, node2vec, which by design did not generate edge type embeddings.

### 3.3 Transductive Inference Testing

The trained models were tested for transductive inference, i.e., unseen link prediction between nodes that were in the training set, on the set-aside test set of edges using embeddings trained with the training set of edges. Metrics are defined later in the paper. To allow us to assess bias (underfitting) and variance (overfitting) of the models, we separately trained the models on all edges (train/validation edges + test edges), and then tested them on the test set. That is, we conducted a separate test on train experiment. As the models can be seen as encode-decoder networks, where the encoder produces embeddings and the decoder uses them to reproduce the graph, transductive inference is an application of the decoder as a scoring function on the embeddings of a given pair of nodes. Mostly, the models supported cosine similarity as the scoring function but HRGAT required ConvKB as the scoring function.

### 3.4 Inductive Inference Method

Inductive inference, as mentioned earlier, is predicting a link between two nodes, where at least one of them was not a part of KGE training. A naïve approach would be to use randomly initialized embeddings for the new node or nodes and use the trained relation type embedding to predict a link between them. Methods used in the general domain are discussed in (Galkin et al. 2022; Li et al. 2023), the latter being a review of the most relevant and latest methods. The general approach involves capturing the structural patterns from the KG during the training, then apply them to learn the representations of new nodes. However, none of them take advantage of the node semantics, for example, as reflected in their attribute values.

Our approach is based on node attribute values, which happened to be text segments in NVKG. The goal here is to estimate KG embeddings of a new node based on its attribute value. The core idea is that using the attribute value of a new node, we identify similar nodes in the attribute value latent space and then from the KG embeddings of the similar nodes, we determine the new node's KG embeddings. Since the node attribute values are text strings in NVKG, we use a latent space of text embeddings. For a new node, using its attribute value text string, we first determine its text embeddings in the textual latent space. Next, we identify its k-nearest neighbors in the textual latent space. Then the weighted average of the neighbors KG embeddings is calculated, using the cosine similarities in the textual latent space as the weights. The result is an estimated KG embedding for the new node, and further reasoning with the new node can be carried out in the KG embedding space. The method is shown in Figure 4 an described in Method 1.0.

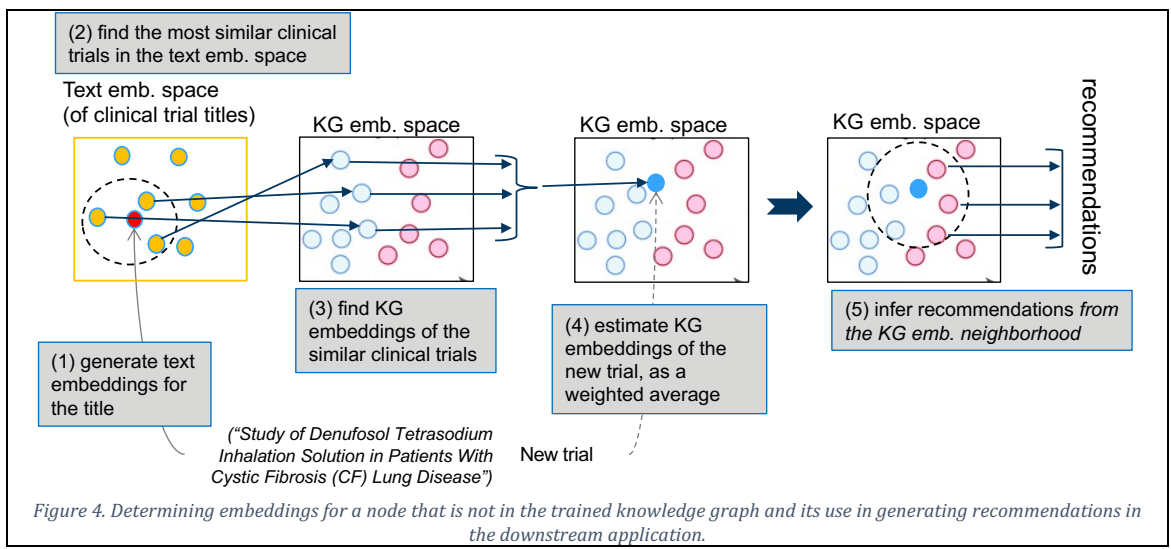

*Figure 4. Determining embeddings for a node that is not in the trained knowledge graph and its use in generating recommendations in the downstream application.*

*Method 1.0: Estimating KGE for a node that is not present in the KGE training set.*

1. $\mathcal{E}_{Txt}$ is the latent space of attribute value strings.
2. $\mathcal{E}_{KG}$ is the latent space of knowledge graph embeddings trained on existing nodes.
3. $N_0$ is the new node with an attribute text value $t_0$. $N_0$ is not in the set of trained KG nodes and therefore has no embeddings in $\mathcal{E}_{KG}$ at this stage
4. $N_1, N_2, \ldots, N_k$ are the k-nearest neighbors of $N_0$ in $\mathcal{E}_{Txt}$ (based on cosine similarity)
5. KG embeddings of $N_0$ in $\mathcal{E}_{KG}$ are then estimated as: $e_{N0} = \frac{1}{k}\sum_k w_i * e_{Ni}$ where $e_{Ni}$ is the embeddings of $N_i$ in $\mathcal{E}_{KG}$ and the weights $w_i$ are the cosine distances between $N_0$ and $N_i$ in $\mathcal{E}_{Txt}$ and therefore are in the range 0 to 1.0.

The embeddings computed for the new node can be used for further inference in the knowledge graph embeddings latent space ($\mathcal{E}_{KG}$). $\mathcal{E}_{Txt}$ was trained using *fasttext* on textual strings of all textual entities in clinical trials of CT.gov. The trained model was then used to determine text embeddings of an $N_i$ from its attribute text value.

### 3.5 Generating Design Recommendations for the Downstream Application

The method for generating design recommendations for a new clinical trial uses KG embeddings estimated for a new node as described in the previous section. Once the KG embeddings are estimated for a new node corresponding to a new clinical trial, we select the $k$ nearest neighbors of the estimated embeddings in the $\mathcal{L}_{KG}$ of the required type as the recommendations, as shown in Figure 4. Note that since the KG latent has multiple node types (e.g., ICR, ECR, pep, sep, and NCT) the k-nearest neighbors of a specific type should be selected (e.g., ICR) to recommend ICRs (inclusion criteria). For the same reason, recommendations of different design elements can be made for the new clinical trial using the same technique. The method is detailed below.

*Method 2.0: Generating design recommendations for a new trial using its (proposed) title.*

1. Given, k (number of recommendations required) and D (the design element type) for an unseen clinical trial C.
2. A clinical trial node of type $N_0$ was created for C with the (proposed) trial title as its attribute string value.
3. KGE for $N_0, e_{N0}$, was estimated using Method 1.0.
4. Then k nearest neighbors of type D were identified in $\mathcal{L}_{KG}$ and their attribute string values were presented to the user as recommendations in their cosine similarity order.

Furthermore, Method 1.0 is a general technique to obtain KG embeddings for any new textual-type entity. For instance, let us assume that the new trial design continues, and the designer formulated a couple of endpoints and a few eligibility criteria. Then, their KG embeddings can also be obtained in the same way as we did for the new clinical trial, i.e., use the textual embeddings to find the k-nearest neighbors and use their weighted average of KG embeddings as the KG embeddings of the newly formulated design elements. (Categorical entities such as AGE and Gender already have KG embeddings for all values and, in the case of UMLS entities, require a separate embeddings space of their own.) These estimated embeddings of a clinical trial can be further used as inputs in a model to predict outcomes such as the probability of completion for the trial being designed.

The estimated embeddings of a new clinical trial element can be used for generating a wide range of design elements such as primary and secondary endpoints and inclusion and exclusion criteria. In addition, other relevant clinical trials can be found. Relevant drugs, targets, and mechanism of action can also be found for a new clinical trial being designed. The k-nearest neighbor search based on the estimated embeddings makes the KG a versatile tool for various downstream applications.

**Recommendations Evaluation Experiment**: To evaluate the relevance of the generated recommendations, we conducted the following experiment. Additional 449 clinical trials from CT.gov were identified using stratified random selection across the disease areas listed in Table 2. This is considered as the "blind" set for the experiment in the sense that none of the data from these trials were a part of the KGE training. From each of these trials and for each of its design element type D, up to k recommendations, were generated using Method 2.0. Then, we determined the text similarity of the recommendations and the actual elements of the blind-test clinical trial using *fasttext*. For example, if a blind-test trial has $j$ inclusion criteria and $k$ recommendations were made, we calculated cosine similarity of $k$ x $j$ pairs, and for each $j$, we recorded the highest similarity score and the position of the recommendation that scored the highest similarity, the first recommendation is counted as 1. We chose primary endpoints and inclusion criteria for the evaluation.

### 3.6 Performance Metrics

For transductive inference, which also establishes the quality of KGE in terms of their ability to reproduce the input KG in the encode-decoder paradigm, we used the standard KG performance metrics, i.e., mean reciprocal rank (MRR) and hits at $k$ (Hits@$k$) (Bordes et al. 2013; Yang et al. 2015; Zheng et al. 2021). The experiment involves calculating rank for each triplet $(h, r, t)$ where $h$ is the head node, $t$ is tail node, and $r$ is relation type between them. Specifically, for each $h$, all tail nodes from the domain of the tail node type, except those in the train and validation sets as well as the rest of the nodes in the test set, were scored using the scoring function of the KGE model (usually, it is the cosine similarity, but not always). The scored tails are ordered in the descending order of the scores. The rank of $t$ for a specific $(h, r, t)$ is the *position* of $t$'s score in this list counting down from the top, the first one being 1. Subsequently, ranks for all h are also determined by fixing t in a similar way. Overall, MRR is reciprocal of all ranks thus calculated. Hits@$k$ is simply the fraction of ranks less than or equal to k. Note that both MRR and Hits@$k$ range from 0 to 1.0, the higher being the better.

For the recommendations, we used mean rank and mean reciprocal rank (MRR) also but defined differently as for the recommender systems. The rank is the position of the highest similarity scoring recommendation in the list of recommendations, with the first one being one. Average of the ranks over all queries is reported as the mean rank and the mean of rank reciprocals is reported as MRR. We also reported the average of the highest similarity scores over all queries.

## 4 RESULTS

### 4.1 Transductive inference performance (as measured by link prediction)

The first set of results show link prediction accuracy on the test dataset, for various KGE methods (see Table 4). The left side of the table is the results for NVKG, and the right side of the table is reproduced from (Zheng et al. 2021) for PharmKG for the comparison purposes. For NVKG, the best MRR and Hits at 3 and 10 were achieved by HRGAT, which is an adaptation of the attention-based transformer, GAT. TransE achieved the second best performance for MRR and Hits at 3 and 10. HRGAT and TransE switched places for Hits@1, where TransE achieved the best performance while HRGAT achieved the second best performance.

In comparison, HRGAT achieved the best MRR and Hits at 1, 3, and 10 for PharmKG, and the semantics based KGE method, ComplEx, achieved the second best performance across the board for the KG. HRGAT performed the best and achieved similar MRR for both PharmKG and NVKG despite the structural differences between the two knowledge graphs. It should be noted that the MRR values were 0.154 and 0.156 for the KGs, indicating that there is room for improvement for both knowledge graphs.

*Table 1. Transductive link predictive for various KGE methods. The best performance is achieved by HRGAT followed by TransE. Results from PharmKG were reproduced here for comparison.*

| Method | NVKG | | | | PharmKG | | | |
|---|---|---|---|---|---|---|---|---|
| | Hits@1 | Hits@3 | Hits@10 | MRR | Hits@1 | Hits@3 | Hits@10 | MRR |
| Node2Vec | 0.018 | 0.063 | 0.163 | 0.068 | - | - | - | - |
| TransE | **0.067** | <u>0.127</u> | <u>0.224</u> | <u>0.119</u> | 0.034 | 0.092 | 0.198 | 0.091 |
| TransR | 0.038 | 0.083 | 0.164 | 0.082 | 0.030 | 0.071 | 0.155 | 0.075 |
| ComplEx | 0.012 | 0.027 | 0.061 | 0.032 | <u>0.046</u> | <u>0.110</u> | <u>0.225</u> | <u>0.107</u> |
| ConvKB | 0.048 | 0.094 | 0.164 | 0.085 | 0.000 | 0.107 | 0.209 | 0.106 |
| HRGAT | <u>0.055</u> | **0.192** | **0.352** | **0.156** | **0.075** | **0.172** | **0.315** | **0.154** |
| NodePiece | 0.001 | 0.002 | 0.004 | 0.003 | - | - | - | - |

The second set of results in Table 5 show link prediction accuracy for test on trained data for NVKG, providing an opportunity to assess the generalizability gap (i.e., bias and variance). The table shows performance for the set-aside test set (replicated from Table 4) for the comparison purposes.

The test-on-train results were substantially higher than their corresponding set-aside test results. HRGAT achieved 0.609 MRR on the test-on-train dataset but only 0.156 on the set-aside dataset. Similarly, HRGAT Hits@1 is an order of magnitude better on test-on-train dataset compared to the set-aside dataset results. ConvKB performance has distinctly improved and achieved the second best MRR and Hits@1 in the test-on-train results.

*Table 2. Test-on-train results and comparison with test on a set-aside dataset.*

| Method | NVKG – train, valid, test split | | | | NVKG – test on train | | | |
|---|---|---|---|---|---|---|---|---|
| | Hits@1 | Hits@3 | Hits@10 | MRR | Hits@1 | Hits@3 | Hits@10 | MRR |
| Node2Vec | 0.018 | 0.063 | 0.163 | 0.068 | 0.257 | 0.597 | 0.723 | 0.419 |
| TransE | **0.067** | <u>0.127</u> | <u>0.224</u> | <u>0.119</u> | 0.247 | <u>0.616</u> | <u>0.764</u> | 0.452 |
| TransR | 0.038 | 0.083 | 0.164 | 0.082 | 0.294 | 0.441 | 0.610 | 0.400 |
| ComplEx | 0.012 | 0.027 | 0.061 | 0.032 | 0.335 | 0.372 | 0.624 | 0.370 |
| ConvKB | 0.048 | 0.094 | 0.164 | 0.085 | <u>0.379</u> | 0.548 | 0.719 | <u>0.493</u> |
| HRGAT | <u>0.055</u> | **0.192** | **0.352** | **0.156** | **0.501** | **0.659** | **0.841** | **0.609** |
| NodePiece | 0.001 | 0.002 | 0.004 | 0.003 | 0.006 | 0.012 | 0.029 | 0.015 |

### 4.2 Inductive inference performance (as measured in the context of the downstream application)

The results for evaluating design recommendations for a new clinical trial under design, using inductive inference, are shown in Table 6. The results correspond to two design elements, primary endpoints and inclusion criteria, in the two horizontal sections of the Table. The two vertical sections of the Table show results for the two KGE methods, namely Node2Vec and HRGAT. As mentioned earlier, for the primary endpoints, results for N = 3, 5, and 10 recommendations were shown, and for the inclusion criteria, results for N = 10, 30, and 50 top recommendations were shown.

We note that the highest MRR and the lowest mean rank were achieved when the fewest recommendations were presented, whereas the highest average similarity to the ground truth (i.e., to the corresponding actual clinical trial recorded in CT.gov) were achieved when the most recommendations were provided.

Both HRGAT and Node2Vec performed about the same in terms of MRR, mean rank, and average similarity for both the design elements. Note that the transductive link prediction results showed HRGAT performing distinctly superior to Node2Vec.

Overall, the semantics-based inference method we proposed here performed well. For the primary endpoints, Node2Vec and HRGAT provided the best recommendation in the mid-position of the list on average, and the best recommendation agreed with the ground truth in the range of 0.70 to 0.78 (depending on the number of recommendations and the KGE method). For the inclusion criteria, the performance of both methods is even better as the best recommendation was in the top 1/3 of the list on average and the average similarity ranged from 0.76 to 0.83.

*Table 3. Recommendations performance*

| Clinical trial elements | Top N recomm. | Node2Vec | | | HRGAT (Transformer GNN) | | |
|---|---|---|---|---|---|---|---|
| | | mean rank | mrr | average similarity | mean rank | mrr | average similarity |
| Primary Endpoints (n=854) | 3 | **1.72** | **0.713** | 0.698 | **1.73** | **0.697** | 0.721 |
| | 5 | 2.33 | 0.602 | 0.722 | 2.62 | 0.513 | 0.751 |
| | 10 | 4.54 | 0.371 | **0.749** | 4.78 | 0.344 | **0.779** |
| Inclusion Criteria (n=3477) | 10 | **3.29** | **0.461** | 0.765 | **3.42** | **0.441** | 0.762 |
| | 30 | 11.3 | 0.190 | 0.815 | 11.9 | 0.174 | 0.812 |
| | 50 | 20.7 | 0.123 | **0.832** | 21.4 | 0.120 | **0.829** |

## 5 DISCUSSION <START>

Several observations can be made based on the results. HRGAT achieves state of the art performance for transductive inference, confirming the advantages of Attention-based transformers for GNN. Performance of KGE methods for NVKG is similar to PharmKG despite the key structural differences in the graph structure. While NVKG has 14 node types, 15 edge types, and only 1-1 node to node relations, PharmKG has three node types, 29 edge types, and multiple node-to-node relations. The link prediction performance of the two knowledge graphs is similar – MRR for HRGAT is 0.156 for NVKG and 0.154 for PharmKG – and the trend is similar for the other KGE methods. One possible explanation is that KGEs are independent of the graph structure. Another explanation is that while the relation type embeddings were well trained, node embeddings are not as well trained in both graphs, resulting in nodes having poor discriminative capability. The inference is likely driven entirely by the type embeddings. The latter view is supported by the low MRR values for all the KGE methods, and relatively smaller differences between the traditional methods (e.g., TransE) and the newer, GNN methods (e.g., HRGAT).

Another observation is that the models do not generalize very well, based on the significant gap between the results from the test-on-train and set-aside test set experiments on NVKG (similar results were not available for other KGs). For example, HRGAT MRR is 0.609 for the test-on-train experiment but only 0.156 on the set-aside experiment. Similar performance gap can be seen for the other KGE methods also. Once again, the possibility is that the node embeddings may not be training well enough to be discriminative and that most of the model training is captured in relation type embeddings. This suggests including node semantics (i.e., node attributes) in training graph embeddings, in addition to the graph structure, may help improve representativeness.

Yet another observation is that the novel inductive inference method and the downstream application we developed using the inductive inference worked well despite the above limitations of the KGE model performance. The reason for it is the combination of two design elements incorporated into the method: (1) The use of the title of the new clinical trial to embed it among the existing clinical trials using title (text) embeddings; (2) Estimation of the new trial's KGE from the nearest neighbors in the text embedding space and then using this estimated KGE and the rest of trained KGEs, to find the nearest neighbors. With this approach, recommendations can be made for any design element not just those we have experimented here.

What about a search engine based approach for help with clinical trial design? While it is possible to use a search engine APIs, for example, from CT.gov, it is a multi-step process with potentially noisy results. The process would have to start with finding relevant trials, collecting required design elements for them, filtering and consolidating, and prioritizing the data collected. It is not only a long and possibly tedious (if done manually) but can also introduce noise and may miss relevant information. Structured databases created from the content in CT.gov are not likely to fair much better. The use of Large Language Models, like ChatGPT, is another possibility and a topic of future research, where the benefits of an AI-based conversational capability must be weighed against the challenges of hallucinations and lack of currency.

The effectiveness of our approach to inductive inference suggests that the use of node semantics may help improve representation of node embeddings and therefore may improve transductive inference as well. It may well complement the training from graph structure.

It should be noted that our indicative inference approach can be used to estimate KG embeddings of a new node of any type. For example, if the trial design process formulate some of the eligibility criteria, using the text of the designed eligibility criteria, their KG embeddings can be estimated using our approach. Once KG embeddings are estimated, they can be used to predict certain characteristics of the trial being designed, such as the probability of completion.

One last observation is that Node2Vec and HRGAT perform equally well (perhaps, Node2Vec may even be seen as the better of the two) in inductive performance (Table 6) despite considerable difference in MRR when tested on train as shown in Table 5 (i.e., 0.609 vs 0.419). This requires further study; however, we believe that this may be related to the earlier observation on poor discriminative capabilities of KG node embeddings.

## 6 SUMMARY AND FUTURE WORK

In this paper, we proposed fist of a kind knowledge graph for representing clinical trials for design recommendations, assessed various methodologies for knowledge graph embeddings of the graph, and used the KGE along with text semantics in a novel inductive inference method for generating recommendations for a new clinical trial design. We showed publicly available data from clinicaltrials.gov can be effectively used for the study. Results showed that the recommendations approach achieved relevancy of 70%-83%, measured as the text similarity to actual clinical trial elements, and that the most relevant recommendation can be found near the top of list. These results add a new data point in using knowledge graphs for biomedical applications by contributing new insights into the promise and challenges of doing so.

For future work, we will explore using the node semantics in training graph embeddings in addition to the graph structure. Specifically, current graph training methods fall into two categories: (1) Message passing architectures that leverage the direct neighbors of a node in training its representation; (2) Broader graph structure of a node in its neighborhood in addition to its direct neighbors such as the patterns of connections and anchor nodes. However, both approaches typically start with random values for the initial embeddings of a node (or the output of a method that started with initial random values). The node attributes are not considered at any stage of KGE training. We propose exploring an approach that incorporates node attributes as an element of the neural network training.

In another direction, we propose to study downstream applications involving predicting outcomes, such as the probability of completion. The selection or design of end points, eligibility criteria, gender, the age group, locations for the study, and other such factors have an influence on the ability to recruit patients and conduct the study in a timely manner. Instead of building one-off models to predict such factors, KGE from NVKG can be used to make such predictions. Our preliminary studies in this direction are promising and will explore it further.

In yet another direction, we will consider how the model can be used to recommend design elements and relevant trials based on drug targets and/or drug mechanism of action as NVKG is first to integrate clinical trials data with drug characteristics. The targets and mechanism of action can also be used as additional filters in generating recommendations or predicting operational aspects of a proposed design. In conclusion, we showed that a clinical trial KG and embeddings generated for it are versatile and can be a basis for developing helpful tools for clinical trial design.

## ACKNOWLEDGMENTS

We gratefully acknowledge the collaboration of Novartis associates who worked on this and other related projects. We especially thank, Iya Khalil, PhD, for her continuous support and visionary leadership in applying AI to drug discovery.

## REFERENCES

Abu-Salih, Bilal, Muhammad AL-Qurishi, Mohammed Alweshah, Mohammad AL-Smadi, Reem Alfayez, and Heba Saadeh. 2023. "Healthcare Knowledge Graph Construction: A Systematic Review of the State-of-the-Art, Open Issues, and Opportunities." *Journal of Big Data* 10 (1): 81. https://doi.org/10.1186/s40537-023-00774-9.

Ali, Mehdi, Max Berrendorf, Charles Tapley Hoyt, Laurent Vermue, Mikhail Galkin, Sahand Sharifzadeh, Asja Fischer, Volker Tresp, and Jens Lehmann. 2022. "Bringing Light Into the Dark: A Large-Scale Evaluation of Knowledge Graph Embedding Models Under a Unified Framework." *{IEEE} Transactions on Pattern Analysis and Machine Intelligence* 44 (12): 8825–45.

Ali, Mehdi, Max Berrendorf, Charles Tapley Hoyt, Laurent Vermue, Sahand Sharifzadeh, Volker Tresp, and Jens Lehmann. 2021. "PyKEEN 1.0: A Python Library for Training and Evaluating Knowledge Graph Embeddings." *Journal of Machine Learning Research* 22 (82): 1–6. http://jmlr.org/papers/v22/20-825.html.

Alshahrani, Mona, Mohammad Asif Khan, Omar Maddouri, Akira R Kinjo, Núria Queralt-Rosinach, and Robert

Hoehndorf. 2017. "Neuro-Symbolic Representation Learning on Biological Knowledge Graphs." *Bioinformatics* 33 (17): 2723–30. https://doi.org/10.1093/bioinformatics/btx275.

Belleau, F, Nolin M-a, N Tourigny, and others. 2008. "Bio2RDF: Towards a Mashup to Build Bioinformatics Knowledge Systems." *Biomed Inform* 41: 706–16.

Bordes, Antoine, Nicolas Usunier, Alberto Garcia-Durán, Jason Weston, and Oksana Yakhnenko. 2013. "Translating Embeddings for Modeling Multi-Relational Data." In *Proceedings of the 26th International Conference on Neural Information Processing Systems - Volume 2*, 2787–2795. NIPS'13. Red Hook, NY, USA: Curran Associates Inc.

Breit, Anna, Simon Ott, Asan Agibetov, and Matthias Samwald. 2020. "OpenBioLink: A Benchmarking Framework for Large-Scale Biomedical Link Prediction." *Bioinformatics* 36 (13): 4097–98. https://doi.org/10.1093/bioinformatics/btaa274.

Cao, Jiahang, Jinyuan Fang, Zaiqiao Meng, and Shangsong Liang. 2022. "Knowledge Graph Embedding: A Survey from the Perspective of Representation Spaces." *ArXiv Preprint ArXiv:2211.03536.* https://arxiv.org/pdf/2211.03536.pdf.

Chandak, Payal, Kexin Huang, and Marinka Zitnik. 2023. "Building a Knowledge Graph to Enable Precision Medicine." *Scientific Data* 10 (1): 67. https://doi.org/10.1038/s41597-023-01960-3.

Chen, Ziqi, Bo Peng, Vassilis N Ioannidis, Mufei Li, George Karypis, and Xia Ning. 2022. "A Knowledge Graph of Clinical Trials (CTKG)." *Scientific Reports* 12.

Dai, Yuanfei, Shiping Wang, and Neal N Xiong. 2020. "A Survey on Knowledge Graph Embedding : Approaches , Applications and Benchmarks," 1–29. https://doi.org/10.3390/electronics9050750.

Dettmers, Tim, Pasquale Minervini, Pontus Stenetorp, and Sebastian Riedel. 2018. "Convolutional 2D Knowledge Graph Embeddings." In *Proceedings of the Thirty-Second AAAI Conference on Artificial Intelligence and Thirtieth Innovative Applications of Artificial Intelligence Conference and Eighth AAAI Symposium on Educational Advances in Artificial Intelligence.* AAAI'18/IAAI'18/EAAI'18. AAAI Press.

Fernández-Torras, Adrià, Miquel Duran-Frigola, Martino Bertoni, Martina Locatelli, and Patrick Aloy. 2022. "Integrating and Formatting Biomedical Data as Pre-Calculated Knowledge Graph Embeddings in the Bioteque." *Nature Communications* 13 (1): 5304. https://doi.org/10.1038/s41467-022-33026-0.

Galkin, Mikhail, Etienne Denis, Jiapeng Wu, and William L Hamilton. 2022. "NodePiece: Compositional and Parameter-Efficient Representations of Large Knowledge Graphs." In *ICLR 2022*, 1–25.

Grover, Aditya, and Jure Leskovec. 2016. "Node2vec: Scalable Feature Learning for Networks." In *Proceedings of the 22nd ACM SIGKDD International Conference on Knowledge Discovery and Data Mining*, 855–864. KDD '16. New York, NY, USA: Association for Computing Machinery. https://doi.org/10.1145/2939672.2939754.

Hamosh, A, A F Scott, J Amberger, and others. 2000. "Online Mendelian Inheritance in Man (OMIM)." *Hum Mutat* 15: 57–61.

Hewett, M, D E Oliver, D L Rubin, and others. 2002. "PharmGKB: The Pharmacogenetics Knowledge Base." *Nucleic Acids Res* 30: 163–65.

Hong, Lixiang, Jinjian Lin, Shuya Li, Fangping Wan, Hui Yang, Tao Jiang, Dan Zhao, and Jianyang Zeng. 2020. "A Novel Machine Learning Framework for Automated Biomedical Relation Extraction from Large-Scale Literature Repositories." *Nature Machine Intelligence* 2 (6): 347–55. https://doi.org/10.1038/s42256-020-0189-y.

Li, Qian, Shafiq Joty, Daling Wang, Shi Feng, Yifei Zhang, and ChengWei Qin. 2023. "Contrastive Learning with Generated Representations for Inductive Knowledge Graph Embedding." In *Findings of the Association for Computational Linguistics: ACL 2023*, 14273–87.

Lin, Yankai, Zhiyuan Liu, Maosong Sun, Yang Liu, and Xuan Zhu. 2015. "Learning Entity and Relation Embeddings for Knowledge Graph Completion." In *Proceedings of the Twenty-Ninth AAAI Conference on Artificial Intelligence*, 2181–2187. AAAI'15. AAAI Press.

Liu, Renming, and Arjun Krishnan. 2021. "PecanPy: A Fast, Efficient and Parallelized Python Implementation of Node2vec." *Bioinformatics* 37 (19): 3377–79. https://doi.org/10.1093/bioinformatics/btab202.

Mohamed, S K, V Nováček, and A Nounu. 2020. "Discovering Protein Drug Targets Using Knowledge Graph Embeddings." *Bioinformatics* 36: 603–10.

Nguyen, Dai Quoc, Tu Dinh Nguyen, Dat Quoc Nguyen, and Dinh Phung. 2018. "A Novel Embedding Model for Knowledge Base Completion Based on Convolutional Neural Network." In *Proceedings of the 2018 Conference of the North {A}merican Chapter of the Association for Computational Linguistics: Human Language Technologies, Volume 2 (Short Papers)*, 327–33. New Orleans, Louisiana: Association for Computational Linguistics. https://doi.org/10.18653/v1/N18-2053.

Nickel, Maximilian, Volker Tresp, and Hans-Peter Kriegel. 2011. "A Three-Way Model for Collective Learning on Multi-Relational Data." In *Proceedings of the 28th International Conference on International Conference on Machine Learning*, 809–816. ICML'11. Madison, WI, USA: Omnipress.

Ou, Mingdong, Peng Cui, Jian Pei, Ziwei Zhang, and Wenwu Zhu. 2016. "Asymmetric Transitivity Preserving Graph Embedding." In *Proceedings of the 22nd ACM SIGKDD International Conference on Knowledge Discovery and Data Mining*, 1105–1114. KDD '16. New York, NY, USA: Association for Computing Machinery. https://doi.org/10.1145/2939672.2939751.

Percha, Bethany, and Russ B Altman. 2018. "A Global Network of Biomedical Relationships Derived from Text." *Bioinformatics* 34 (15): 2614–24. https://doi.org/10.1093/bioinformatics/bty114.

Ren, Zhong-Hao, Zhu-Hong You, Chang-Qing Yu, Li-Ping Li, Yong-Jian Guan, Lu-Xiang Guo, and Jie Pan. 2022. "A Biomedical Knowledge Graph-Based Method for Drug–Drug Interactions Prediction through Combining Local and Global Features with Deep Neural Networks." *Briefings in Bioinformatics* 23 (5): bbac363. https://doi.org/10.1093/bib/bbac363.

Santos, Alberto, Ana R Colaço, Annelaura B Nielsen, Lili Niu, Maximilian Strauss, Philipp E Geyer, Fabian Coscia, et al. 2022. "A Knowledge Graph to Interpret Clinical Proteomics Data." *Nature Biotechnology* 40 (5): 692–702. https://doi.org/10.1038/s41587-021-01145-6.

Schlichtkrull, Michael, Thomas N Kipf, Peter Bloem, Rianne van\ den Berg, Ivan Titov, and Max Welling. 2018. "Modeling Relational Data with Graph Convolutional Networks." In *The Semantic Web: 15th International Conference, ESWC 2018, Heraklion, Crete, Greece, June 3–7, 2018, Proceedings*, 593–607. Berlin, Heidelberg: Springer-Verlag. https://doi.org/10.1007/978-3-319-93417-4_38.

Scott, Evans R. 2010. "Fundamentals of Clinical Trial Design." *Journal of Experimental Stroke & Translatin Medicine* 3 (1): 19–27.

Sosa, D N, A Derry, Margaret Guo, Eric Wei, Connor Brinton, and Russ B. Altman. 2020. "A Literature-Based Knowledge Graph Embedding Method for Identifying Drug Repurposing Opportunities in Rare Diseases." In *Pacific Symposium on Biocomputing*, 463–74.

Teru, Komal K, Etienne G Denis, and William L Hamilton. 2020. "Inductive Relation Prediction by Subgraph Reasoning." In *Proceedings of the 37th International Conference on Machine Learning*. Vienna, Austria.

Trouillon, Théo, Johannes Welbl, Sebastian Riedel, Éric Gaussier, and Guillaume Bouchard. 2016. "Complex Embeddings for Simple Link Prediction." In *Proceedings of the 33rd International Conference on International Conference on Machine Learning - Volume 48*, 2071–2080. ICML'16. JMLR.org.

Velickovic, Petar, Guillem Cucurull, Arantxa Casanova, Adriana Romero, Pietro Lio, and Yoshua Bengio. 2018. "Graph Attention Networks." In *ICLR 2018*.

Wang, Q, Z Mao, B Wang, and others. 2017. "Knowledge Graph Embedding: A Survey of Approaches and Applications." *IEEE Trans Knowl Data Eng* 29: 2724–43.

Wang, Siyuan, Zhongyu Wei, Meng Han, Zhihao Fan, Haijun Shan, Qi Zhang, and Xuanjing Huang. 2023. "Query Structure Modeling for Inductive Logical Reasoning Over Knowledge Graphs." In *Proceedings of the 61st Annual Meeting of the Association for Computational Linguistics, Vol1: Long Papers*, 4706–18.

Wang, Yunxia, Song Zhang, Fengcheng Li, Ying Zhou, Ying Zhang, Zhengwen Wang, Runyuan Zhang, et al. 2020. "Therapeutic Target Database 2020: Enriched Resource for Facilitating Research and Early Development of Targeted Therapeutics." *Nucleic Acids Research* 48 (D1): D1031–41. https://doi.org/10.1093/nar/gkz981.

Wishart, D S, Y D Feunang, A C Guo, and others. 2018. "DrugBank 5.0: A Major Update to the DrugBank Database for 2018." *Nucleic Acids Res* 46: D1074-82.

Yang, Bishan, Wen tau Yih, Xiaodong He, Jianfeng Gao, and Li Deng. 2015. "Embedding Entities and Relations for Learning and Inference in Knowledge Bases." *3rd International Conference on Learning Representations, ICLR 2015*.

Zeng, Xiangxiang, Xinqi Tu, Yuansheng Liu, Xiangzheng Fu, and Yansen Su. 2022. "Toward Better Drug Discovery with Knowledge Graph." *Current Opinion in Structural Biology* 72: 114–26. https://doi.org/https://doi.org/10.1016/j.sbi.2021.09.003.

Zheng, Shuangjia, Jiahua Rao, Ying Song, Jixian Zhang, Xianglu Xiao, Evandro Fei Fang, Yuedong Yang, and Zhangming Niu. 2021. "PharmKG: A Dedicated Knowledge Graph Benchmark for Biomedical Data Mining." *Briefings in Bioinformatics* 22 (4).

Zhou, Jie, Ganqu Cui, Shengding Hu, Zhengyan Zhang, Cheng Yang, Zhiyuan Liu, Lifeng Wang, Changcheng Li, and Maosong Sun. 2020. "Graph Neural Networks: A Review of Methods and Applications." *AI Open* 1: 57–81. https://doi.org/https://doi.org/10.1016/j.aiopen.2021.01.001.